\documentclass{article}

\usepackage{microtype}
\usepackage{graphicx}
\usepackage{subfigure}
\usepackage{booktabs} % for professional tables

\usepackage{hyperref}
\usepackage{algorithmic}
\usepackage{multirow}
\renewcommand{\algorithmicrequire}{\textbf{Input:}} 
\renewcommand{\algorithmicensure}{\textbf{Output:}}

\usepackage[accepted]{icml2025}

\usepackage{amsmath}
\usepackage{amssymb}
\usepackage{mathtools}
\usepackage{amsthm}

\usepackage[capitalize,noabbrev]{cleveref}

\theoremstyle{plain}

\theoremstyle{definition}

\theoremstyle{remark}

\usepackage[textsize=tiny]{todonotes}

\icmltitlerunning{Submission and Formatting Instructions for ICML 2025}

\begin{document}

\twocolumn[
\icmltitle{Physics-inspired Energy Transition Neural Network for Sequence Learning}



\icmlsetsymbol{equal}{*}

\begin{icmlauthorlist}
\icmlauthor{Zhou Wu}{a,c}
\icmlauthor{Junyi An}{a,c}
\icmlauthor{Baile Xu}{b,c}
\icmlauthor{Furao Shen}{b,c}
\icmlauthor{Jian Zhao}{d,c}

\end{icmlauthorlist}

\icmlaffiliation{a}{Department of Computer Science and Technology, Nanjing University, Nanjing 210023, China}
\icmlaffiliation{b}{School of Artificial Intelligence, Nanjing University, Nanjing 210023, China}
\icmlaffiliation{c}{State Key Laboratory for Novel Software Technology}
\icmlaffiliation{d}{School of Electronic Science and Engineering, Nanjing University, Nanjing 210023, China}

\icmlcorrespondingauthor{Furao shen}{frshen@nju.edu.cn}
%\icmlcorrespondingauthor{Firstname2 Lastname2}{first2.last2@www.uk}

\icmlkeywords{Machine Learning, ICML}

\vskip 0.3in
]



\printAffiliationsAndNotice{} % otherwise use the standard text.

\setlength{\abovedisplayskip}{4pt}
\setlength{\abovedisplayshortskip}{1pt}
\setlength{\belowdisplayskip}{4pt}
\setlength{\belowdisplayshortskip}{1pt}
\setlength{\jot}{3pt}
\setlength{\textfloatsep}{6pt}	
\setlength{\abovecaptionskip}{1pt}
\setlength{\belowcaptionskip}{1pt}
\setlength{\tabcolsep}{4pt} 

\begin{abstract}
Recently, the superior performance of Transformers has made them a more robust and scalable solution for sequence modeling than traditional recurrent neural networks (RNNs). However, the effectiveness of Transformer in capturing long-term dependencies is primarily attributed to their comprehensive pair-modeling process rather than inherent inductive biases toward sequence semantics. In this study, we explore the capabilities of pure RNNs and reassess their long-term learning mechanisms. Inspired by the physics energy transition models that track energy changes over time, we propose a effective recurrent structure called the \emph{``Physics-inspired Energy Transition Neural Network" (PETNN)}. We demonstrate that PETNN's memory mechanism effectively stores information over long-term dependencies. Experimental results indicate that PETNN outperforms transformer-based methods across various sequence tasks. Furthermore, owing to its recurrent nature, PETNN exhibits significantly lower complexity. Our study presents an optimal foundational recurrent architecture and highlights the potential for developing effective recurrent neural networks in fields currently dominated by Transformer. 
\end{abstract}
%%%% image is also the format of sequence series data

\section{Introduction}
Recurrent Neural Networks (RNNs), a sophisticated architecture for modeling sequences, have successfully addressed numerous challenges in time series prediction~\cite{timeseries}, machine translation~\cite{translation}, and other areas. Although Transformer models have outperformed RNNs in many areas over the past few years, RNNs remain a valuable research direction due to their inherent inductive biases, which align well with the structure of many real-world sequence data.

% However, the ability of these models to process sequence data and their applicability to sequence generation tasks still necessitate continuous in-depth research.

For modeling sequences, RNNs can leverage preceding data to influence succeeding data, making them highly suitable for sequence data tasks. In addition, the model uses the same parameters at all time steps, which greatly reduces the complexity and computational cost. However, this structure also faces challenges: when dealing with long sequences, RNNs suffer from gradient vanishing problems, making it difficult to deal with long-term dependency problem~\cite{gradient_long_term}. To address it, researchers have proposed variants of RNN, such as LSTM~\cite{LSTM}, and GRU~\cite{GRU}, which alleviate the long-term dependency problem by gating mechanism. Nevertheless, these methods are still limited in complex and long sequence tasks. Later, Transformer with self-attention~\cite{transformer} was introduced, enabling the model to simultaneously attend to all positions in the sequence. This effectively addresses the long-term dependency problem but also introduces higher complexity and computational burden.

The core challenge in sequence modeling lies in effectively identifying the dependencies between tokens and retaining the essential information from the antecedent sequence. Currently, sequence models capture long-term dependencies in two primary ways:  through direct modeling of dependencies, which is common in RNN-based models, and incorporating memory mechanisms, as seen in models like LSTM and GRU. While RNN-based models, due to their simpler structure, have lower time complexity, they rely heavily on memory, leading to the well-known issue of forgetting past information. LSTM and GRU address this by introducing dedicated memory and forget mechanisms, improving their ability to capture long-term dependencies. However, both models still rely on sigmoid-based gates for controlling memory, which can lead to information leakage over long sequences. This limitation makes them less effective compared to Transformer models, which capture long-range dependencies by modeling all pairwise interactions within the sequence.

In this paper, we propose \textbf{Physical-inspired Energy Transition Neural Network (PETNN)}, which is a novel recurrent structure for effective sequence modeling. Initially, we observed that the energy transition model closely resembles the memory cell, since the energy change is a time function. This observation inspired the core idea of PETNN, i.e., using energy as the cell state for updates. To address this, we propose a novel memory mechanism in PETNN. Unlike LSTM, which relies on predefined memory and forget gates, PETNN empowers the neurons to autonomously determine what information to learn and update based on the energy state, a process we refer to as the \emph{self-selective information mixing method}. This approach offers two key advantages: first, it allows neurons to dynamically control the proportion of information updates; second, it enables neurons to decide the storage duration of relevant information. 
%To that end, we design a novel memory mechanism. Unlike LSTM, PETNN allows the neuron itself to determine the information to be learned and updated through the energy state, which we call the \emph{self-selective information mixing method}. The advantage of this method is that the proportion of information updates and the storage duration are determined by the neurons themselves.

Empirical evaluations of the PETNN model have been conducted across diverse domains, for example, sequence tasks, which have been monopoly by Transformer-based models. The results show that PETNN can compete with Transformer in a range of tasks, which demonstrates PETNN has great potential to be a strong alternative or complement to Transformer. Moreover, in image classification task, the results show that PETNN can be used not only for recurrent structures, but can also be easily embedded in general neural networks to solve non-sequential problems, laying a solid foundation for our future work. Our contributions are as follows:
% \begin{itemize}
%     \item We propose PETNN, a novel model that enhances long-term dependency handling in RNNs.
%     \item PETNN is not only applicable to time-series tasks, but also be fit for a more general neural network structure to solve tasks for sequence.
%     \item Results show that PETNN breaks the dominance of Transformer in many sequence tasks. In addition, the ablation experiments verify that the memory mechanism in PETNN is effective, which provides a novel insight for future sequence learning research.
% \end{itemize}
% \An{
\begin{itemize}
    \item We propose PETNN, a recurrent neural networks that incorporates wisdom from physics-based spatiotemporal model (energy transition model).
    \item We analyze PETNN's structure and demonstrate its advantages in handling long-term dependencies.
    \item Results show that PETNN outperforms Transformers on several sequence tasks. Ablation experiments confirm the effectiveness of PETNN's memory mechanism, providing new insights for sequence learning research.
\end{itemize}
% }

\section{Preliminary}
In this section, we provide a brief review of the energy transition mechanism~\cite{transition}, which is the prototype for PETNN. Following this, we introduce how to integrate the physical model into the machine learning model.

\begin{figure}[t]
\vskip 0.2in
\begin{center}
\centerline{\includegraphics[width=\columnwidth]{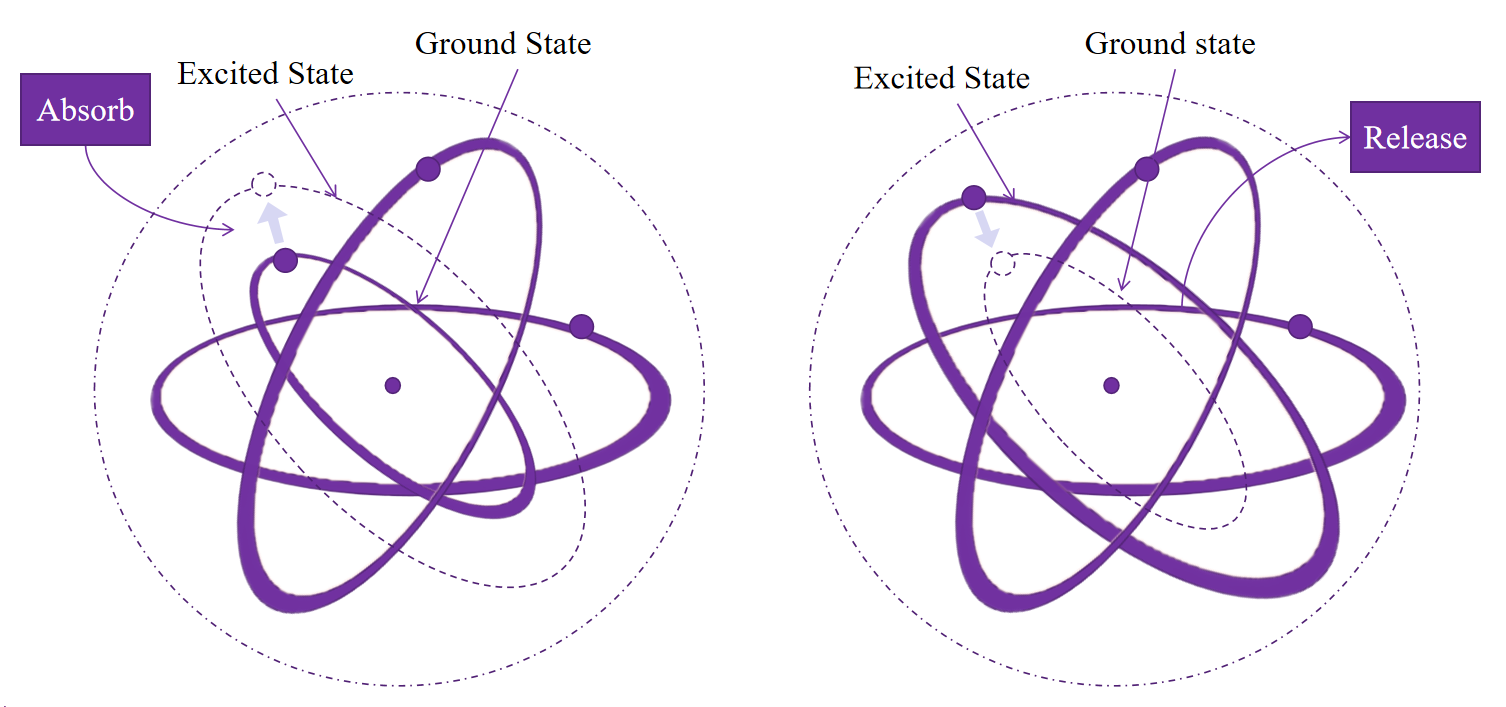}}
    \caption{Quantum theory explains the energy transition model through the processes of absorption and release. }
    \label{releasing and emission}
\end{center}
\vskip -0.2in
\end{figure}

In physics, atoms are characterized by discrete energy levels. When an external energy source interacts with an atom in its ground state, the atom absorbs energy and enters an excited state. However, this excited state is unstable, and the atom eventually releases the absorbed energy after a period of time~\cite{light}, returning to a lower energy state. This process, as shown in Figure~\ref{releasing and emission}, widely applied in quantum physics, is exemplified by the \textbf{photoelectric effect} ~\cite{photoeletric}. More details on the physics of energy transition can be found in Appendix A. Moreover, we refer interested readers to ~\cite{Griffiths}.

Inspired by this theory, our model captures and retains informative features in a way that resembles energy transition model theory. Just as excited states in energy transition models are deterministic, depending on the system's properties and external input, each basic unit in our model adjusts its lifetime based on specific needs.

During this retention period, information is accumulated in the mixed memory, similar to how energy input triggers a transition to an excited state in quantum systems, resulting in an elevated information state.

The information updating process resembles the collapse of the quantum wave function\cite{collapse}, where the updating function deterministically converts the mixed state into a specific state when the lifetime ends. 

After this period, part of information is discarded, akin to entropy reduction~\cite{entropy}, in which redundant or irrelevant information is removed to maintain system efficiency and low entropy.

Specifically, unlike the discrete energy levels in quantum systems, our model uses thresholds to distinguish between two states while processing continuous inputs, enabling dynamic and flexible adaptation. This design ensures that our model learns effectively from the data flow, while maintaining its foundational similarity to the energy transition model, with a focus on dynamically managing information storage and forgetting.
% During this retention period, information is accumulated in the mixed memory, as is described \textbf{excited state being the result of energy accumulation} in quantum physics. 

\section{Model}
% \An{In this section, we propose a novel recurrent structure, which we term the ``Quantum Neural Network (PETN)'', that draw the insight in physics model.}
In this section, we introduce a novel neural network architecture named \emph{Physics-inspired Energy Transition Neural Network (PETNN)}. We begin by outlining the foundational neuron structure, which directly derives insights from energy transition models. Subsequently, we leverage PETNN to build intricate neural networks. Finally, we provide a comprehensive discussion on its physical inspiration

% The overview of our neurons is shown in figure \ref{the basic model of the quantum neurons}. 

% In this section, we will first briefly introduce the basic physical theory of energy absorption and energy level transition, then introduce the basic PETN neurons and their bionic significance. Then we make a simple analysis of PETN and discuss why PETN has stronger mathematical expressiveness than traditional neuron modules. Finally, we use basic neurons to construct neural networks.\\
\begin{figure}[t]
\vskip 0.2in
\begin{center}
\centerline{\includegraphics[width=\columnwidth]{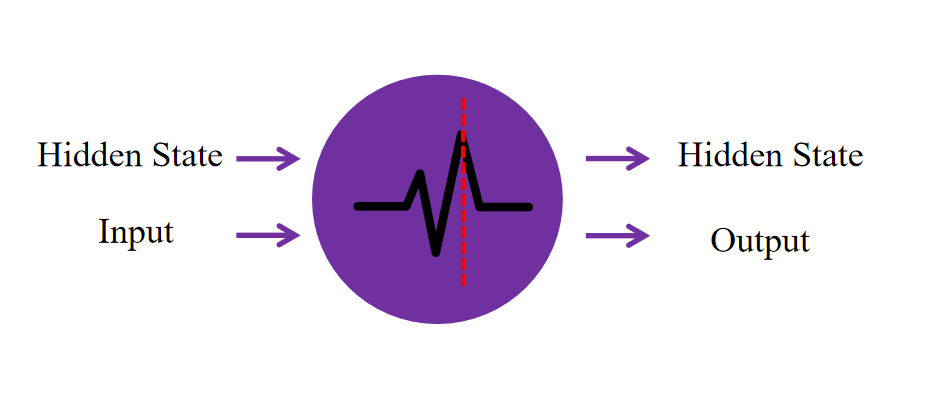}}
    \caption{A basic neuron in PETNN. The black waveform represents stored neuronal information over time. When the signal reaches the red dashed line, the neuron releases this information, returning to ground state.  }
    \label{the basic model of the quantum neurons}
\end{center}
\vskip -0.2in
\end{figure}

\subsection{Physics-inspired Energy Transition Neuron}

Inspired by the energy transition model, we design a neuron structure. As shown in Figure~\ref{the basic model of the quantum neurons}, the neuron can be regarded as the atom. We use a series of variables to mathematically represent it, which are defined as follows.

\begin{itemize}
    \item \textbf{Remaining Time $\mathbf{T}_{t}$}: the remaining time of cell state at step t, corresponds to the residence time of atoms in excited state.
    \item \textbf{Cell State $\mathbf{C}_{t}$}: the state of the neuron at step t, which can be regarded as the energy state of the neuron.
    \item \textbf{Hidden State~$\mathbf{S}_{t}$}: the memory in time step t, as another input of the neuron at time step t+1 except input $X_{t+1}$.
\end{itemize}
%  which is the same as the excited state lifetime in physics。

\begin{figure}[t]
\vskip 0.2in
\begin{center}
\centerline{\includegraphics[width=\columnwidth]{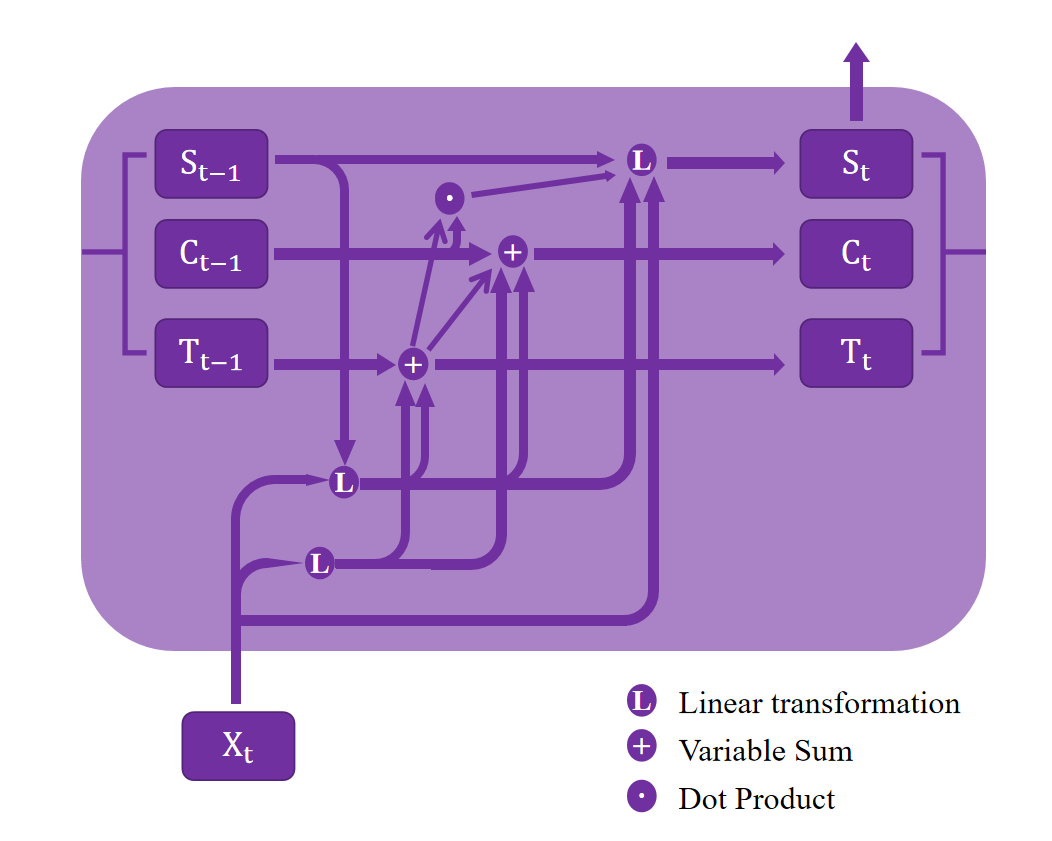}}
    \caption{Internal architecture of PETNN.}
    \label{the inner construction of the quantum neurons}
\end{center}
\vskip -0.2in
\end{figure}

In the beginning, the cell state of the neuron can be regarded as ground state. When the neuron receives external influences, i.e. input signals ($\mathbf{X}_{t}$ and $\mathbf{S}_{t-1}$), the cell state is updated. This detailed computational process is illustrated in Figure~\ref{the inner construction of the quantum neurons}. Similar to the energy transition model, the cell state selectively absorbs information from the input and clears it when the remaining time $T_t$ reaches $0$. Though adjusting the remaining time $T_t$, the neuron can carry complete history information to any subsequent time step. When we build a neural network based on such neurons, the model can distinguish through important information and deliver it to distant time steps.

%  by absorbing the input and last state, 

\subsection{Update process in PETNN}

After defining the basic neuron, we use the input signal to update its state, which is shown in Figure~\ref{the inner construction of the quantum neurons}. In order to express this process clearly, we first define some intermediate variables as follows.

\begin{itemize}
    \item \textbf{Time increment~$Z_t$}: used to update the remaining time.
    \item \textbf{Energy injection~$Z_c$}: used to update the cell state.
    \item \textbf{State update weight~$Z_w$}: used to update the hidden state.
\end{itemize}

Both the input and the previous state of the neuron jointly influence the neuron’s next time step, as well as its residence time and energy absorption. Therefore, we concatenate these two variables: input $X_t$ and the hidden state $S_{t-1}$, and transform them through three separate linear transformations into. The process can be formulated as
\begin{equation}
    \begin{array}{l}
        Z_t = W_{Z_t}\cdot[X_t,S_{t-1}] + b_{Z_t} \\
        Z_c = W_{Z_c}\cdot[X_t,S_{t-1}] + b_{Z_c} \\
        Z_w = W_{Z_w}\cdot[X_t,S_{t-1}] + b_{Z_w} . 
    \end{array}
    \label{linear_transformer_1}
\end{equation}

Meanwhile, to make the model closer to the physics theory, we introduced the scale parameter and the bias parameter:
\begin{itemize}
    \item \textbf{Time decay rate~(scale parameter)~$R_t$}: time measurements corrected factor.
    \item \textbf{Ground state level~(bias parameter)~$I_t$}: the initial value after energy reset.
\end{itemize}
these are the results of the linear transformation of the input,
\begin{equation}
    \begin{array}{l}
           I_t = W_{I_t}\cdot X + b_{I_t}\\
           R_t = W_{R_t}\cdot X + b_{R_t} ~.
    \end{array}
    \label{linear_transformer_2}
\end{equation}
% For the input, we calculate the required variables according to the above processes. 
After defining all the variables, we come to the updating process. First of all, we update the remaining time by simple addition and coefficient multiplication. Next, the energy is updated according to the remaining time. If the remaining time $\mathbf{T}_{t}$ is less than zero, the current energy is released and returned to the ground state, which can be controlled by binary variable $\mathbf{m}$, and then the energy is updated. Otherwise, the update is performed normally. After all the above has been done, we come to the final process, the update of the hidden state, we use the self-selective information mixing method to update all the information neurons learned. Formally, all the updating processes are
\begin{equation}
    \begin{array}{l}
           T_t = R_t \cdot \sigma(T_{t-1}+Z_t)-1 \\
           C_t = (1-m) \cdot C_{t-1} + m \cdot I_t + Z_c\\
           h_t = f(S_{t-1},(1-m)\cdot C_{t-1},X_t) \\
           S_t = \sigma\big((1-Z_w)\cdot S_{t-1}+Z_w \cdot h_{t}\big) ~,
    \end{array}
    \label{update}
\end{equation}
where $\mathbf{h_{t}}$ represents the information learned at time step t, which is derived based on the state from the previous time step and the input $\mathbf{X_t}$, m is the switch of the release, $f$ denotes the set of mapping applied to the input and previous state to compute $\mathbf{h_{t}}$.

All the above is the overview. Then, we will describe the memory cell in detail.

\subsection{Memory Cell}
% Recall that the remaining time, 
\paragraph{Remaining Time} The time variable is introduced for each neuron. It is primarily determined by the characteristics of the input and the current state of the neuron. Furthermore, to more accurately characterize the temporal variation, we additionally introduce the time decay factor to take into account relativistic effects, which allows for a more precise representation within the physical system, this process can be written as
\begin{equation}
    T_t=R_t\cdot \sigma(T_{t-1}+Z_t) -1 ~,\\
    \label{time_equation}
\end{equation}
where $R_t$ denotes the time decay factor, $Z_t$ denotes the time increment and $\sigma(\cdot)$ represents the activation function.

For ease, we define the time switch variable $\mathbf{m}$, to determine whether we need to release energy at this time when the value is below 0. At this point, we set~$\mathbf{m}$ to 0, if $T_t \leq 0$, otherwise 1, then the mathematical expression is
% todo 修改公式
\begin{equation}
    m = \left .\{
 \begin{array}{ll}
     1 & if ~T_t < 0\\
     0 & if ~T_t \geq 0~.
 \end{array} 
 \label{time threshold}
   \right.
\end{equation}

\paragraph{Cell State} The neuron acts as a conservation of energy, and the excited state lifetime of energy in the neuron serves as a switch for the release of information. When the remaining time $T_t$ reaches a predetermined threshold, the system performs an energy release operation, whose mathematical formula is
\begin{equation}
    C_t = (1 - m)\cdot C_{t-1}~,
    \label{initial time equation}
\end{equation}
where $C_{t-1}$ refers to the cell state in the last time step.

In quantum physics, the ground state is of great significance. Given that the ground state energy is inherently linked to the state of the atomic system, its role in information encoding and processing is of particular importance, then we can transform the Equation ~\ref{initial time equation} to Equation~\ref{final time equation},
\begin{equation}
    C_t = (1 - m)\cdot C_{t-1} + m\cdot I_t ~,
    \label{final time equation}
\end{equation}
where $I_t$ indicates the ground state of the energy level, if $m$ = 1, the energy state will return to the ground state and releases redundant information. At each time-step, the constant process absorb input energy and update the state of the neuron no matter whether the transition happen or not, so we have:
\begin{equation}
    C_t = (1 - m)\cdot C_{t-1} + m \cdot I_t + Z_c~.
    \label{cell state}
\end{equation}
% To prevent issues such as gradient explosion and vanishing gradients, we introduce an activation function before the current state.
%Next, we model the state of the atom itself. We consider the energy carried by the input as the information to learn. 
\paragraph{Hidden State}
After that, we simulate the state of an atom using the hidden state of a neuron. Analogy to the fundamental principles of physics, the neuron's status is decided by its previous state and the energy of the input. Consequently, the intermediate status expression can be formulated as
\begin{equation}
    h_{state} = \sigma (W_h\cdot [X,S_{t-1}\cdot (1-m)\cdot C_{t-1}]+b_h)~,
    \label{state definition}
\end{equation}
where $S_{t-1}$ denotes the hidden state of the last step, $(1-m)\cdot C_{t-1}$ denotes whether the energy has been released, and $\sigma(\cdot)$ represents the activation function.

To update the hidden state, we propose a self-selective information mixing method, inspired by the neuroscience. This method mimics the self-regulatory process of selective mixing and updating, enabling enhanced dynamic learning capabilities and environmental adaptability. Detailed formulation are as follows:
\begin{equation}
     S_t=\sigma \big((1-Z_w)\cdot S_{t-1}+Z_w \cdot h_{state}) \big)~,
    \label{update}
\end{equation}
where $Z_w$ represent the recognition bias of content in now and the last state, which offers a novel perspective on optimizing autonomous information storage structures in changing environments. To facilitate better adaptation to downstream tasks, such as prediction, a gating function may be introduced outside the hidden state.
%where $h_{state}$ is the information lerned in the current time step.

%\paragraph{Output} The above are the core component of PETNN. Next, we will process the outputs of the neurons. Following the principles of RNNs, we compute the output at step t based on hidden state $S_t$, which can be written as
%\begin{equation}
%    O_t = g(S_t)~,
%5\end{equation}
%where $g(\cdot)$ function represents the fully-connected layer with Relu activation, and the output can be used for downstream tasks such as prediction and generation.

%In this section, we will discussion why the internal mechanism will work as the physics system and why the system can remember the useful information.

By drawing an analogy from the energy level transition process in physics and self-cognitive bias in neuroscience, we complete the information accumulation and update process. This approach allows the model to dynamically adapt to new inputs while retaining relevant past knowledge. Integrating these interdisciplinary principles strengthens the model's theoretical framework making it more effective in dynamic environments. This combination of concepts not only validates the model’s effectiveness but also offers new perspectives on information processing in artificial systems.

Above all, the main components of PETNN are described here, with additional details provided in Appendix B.
%based on the physics and neuroscience, we complete the information accumulation and update process make analogy from the energy level transition process and self-cognition bias, which give a more solid theoretical foundation and a new perspective for our model works.

%The core of \textbf{PETNN} is grounded in the principles of physics and neuroscience, providing a solid theoretical foundation. 

% Finally, we use this updating process in Algorithm~\ref{Algorithm}, as shown in Appendix B, to build the neural network. In particular, all the variables in time step 0 are set to zero, and the progress starts in $t=1$. In this paper, we build our model based on the RNN paradigm.

%\begin{figure}[t]
%    \centering
%    \includegraphics[width=\columnwidth]{pic/neural_network.png}
%    \caption{An overview of PETNN. PETNN was constructed in the form of RNN. }
%    \label{Neural network structure constructed using quantum neurons}
% \end{figure}
%%%%
% \subsection{Theoretical Analysis}
% In this section, we will do some extra justification on how the PETNN corresponding to the physics
\subsection{Physics-Inspired Design Rationale}
\setlength{\parskip}{0.5 ex}
The architecture of PETNN is rigorously derived from first principles of quantum energy transition theory, establishing direct mathematical mappings between physical laws and model components. This ensures that the model’s core mechanism operates under physics-driven constraints while maintaining computational tractability.

\paragraph{Energy Transition as a State Update Rule}
The cell state ($C_t$) in Equation~\ref{cell state} emulates discrete energy transitions in quantum systems. Key components include:
\begin{itemize}
    \item {Energy Injection (\( Z_c \))}: Analogous to photon absorption, \( Z_c \) quantifies input-driven energy accumulation.  
    \item {Ground State Reset (\( m \cdot I_t \)):} When \( T_t < 0 \), the cell state is reset to a learnable ground level \( I_t \), ensuring adaptive information retention.
\end{itemize}
This formulation leverages the concept of energy dynamics: Information retention is proportionally regulated by input energy levels, replacing heuristic gating with physics-inspired adaptive control.

\paragraph{Time Decay as a Relaxation Process}
In quantum mechanics, the spontaneous decay of an excited state is governed by the \textbf{relaxation time \( \tau \)}, defined via the rate equation:  
\begin{equation}
    \frac{dN}{dt} = -\frac{1}{\tau} N \quad \Rightarrow \quad N(t) = N_0 e^{-t/\tau},
    \label{relaxation}
\end{equation}
where \( N(t) \) represents the population of the excited state.  

In PETNN, we model the remaining time \( T_t \) as a \textbf{discretized analog} of this process. First, we define the continuous-time dynamics of \( T(t) \):  
\begin{equation}
  \frac{dT}{dt} = -\frac{1}{\tau} T + Z_c(t),
  \label{derived_formation}
\end{equation}

where \( \tau \) is a learnable parameter loosely inspired by quantum relaxation time, and \( Z_c(t) \) is the energy injection term balancing physics-inspired dynamics with data-driven adaptability

Then discretizing via the Euler method with step size \( \Delta t = 1 \) we will have:  
\begin{equation}
    T_{t} = T_{t-1} + \Delta t \left( -\frac{1}{\tau} T_{t-1} + Z_t \right) = \left(1 - \frac{1}{\tau}\right) T_{t-1} + Z_t.
\end{equation}

To align with the activation function \( \sigma(\cdot) \) and ensure bounded outputs, we refine this as:  
\begin{equation}
    T_t = R_t \cdot \sigma(T_{t-1} + Z_t) - 1,  
\end{equation}
where \( R_t = \frac{1}{\tau} \) explicitly links the time decay rate to the quantum relaxation time.

\section{Related Work}
%, manufacturing~\cite{manufacture}, and healthcare~\cite{health}
Time series analysis is of great significance in finance~\cite{finance}, economics~\cite{economics}, meteorology~\cite{meterology} and so on, which identifies trends, cycles, and seasonal patterns to aid decision-making and forecast future values. However, it also faces challenges such as complex trends, non-stationary data, long-term dependencies, multivariate interdependence, missing values, noise, and real-time processing.Selecting suitable models and conducting thorough evaluation are crucial for ensuring prediction accuracy and reliability.

Traditional statistical methods, such as ARIMA~\cite{ARIMA} and Holt-Winters~\cite{Holt-Winters}, rely on manually extracted rules for predictions, but they struggle to handle the complexity of real-world time series. With the development of machine learning, researchers have proposed MLP-based methods. Although these methods outperform traditional approaches to some extent, their effectiveness is limited as they fail to capture the sequential dependencies inherent in time series.

Subsequently, RNN-based models were developed to capture the temporal continuity of time series through recurrent structures. However, they are prone to gradient explosion and vanishing gradient problems, which hinder the capture of long-term dependencies. LSTM and GRU have significantly addressed these issues, laying the groundwork for further advancements in deep learning. Nonetheless, their reliance on the forgetting mechanism can still lead to the loss of crucial long-term dependencies. Quasi-RNN~\cite{quasi} is the latter variant that combines the strengths of both CNNs and RNNs. It captures local dependencies through convolutional operations along the temporal dimension, while leveraging recurrent connections to effectively process sequential data. However, the active selective strategy proposed by us shows another insight.

Transformer-based methods have shown strong performance. By incorporating the self-attention mechanism, they effectively capture the relationships between time steps, enabling the learning of long-term dependencies. Furthermore, Autoformer~\cite{AUTOFORMER} improves prediction accuracy through the automatic correlation mechanism; Informer~\cite{informer} uses sparse self-attention mechanism and spatial scaling technology to improve prediction efficiency and accuracy; ETSformer~\cite{etsformer} combines the transformer framework and the classic exponential smoothing method to handle long-term dependencies and seasonal changes. Although Transformer-based methods excel in many areas, their high computational overhead remains a significant drawback. 

% In contrast, the model we proposed in this paper requires a relatively small amount of computation, while still able to achieve ideal results.
In contrast, the model we propose strike a balance between computational burden and performance. Unlike the currently popular Physics-Informed Neural Networks (PINNs), PETNN is based on a heuristic modeling approach grounded in physical principles. While PINNs incorporate physical constraints into neural networks to solve partial differential equations (PDEs) and ordinary differential equations (ODEs), effectively addressing computational challenges in high-dimensional spaces and complex boundary conditions~\cite{PINN}, PETNNs, starting from energy transition models, focus on the foundational aspects of neural network modeling. By integrating physical insights and multidisciplinary methods, PETNNs offer a novel perspective on multi-task sequence modeling.

%% image classification in appendix
\section{Experiments}
In this section, we compare the performance of PETNN with traditional baseline models across three tasks. To ensure a fair comparison, all models are evaluated under the same framework. The evaluation metrics include accuracy, MSE, and MAE. We assess the models on tasks such as time series forecasting and text sentiment classification. Additional details regarding the experimental setup, model configurations, and the image classification task used for further validation can be found in Appendix C.

\begin{table*}[t]
\caption{Long-term forecasting tasks on basic models. The past sequence is set as 96 for all datasets. All the results are average from 4 different prediction lengths, that is \{96,192,336,720\}. The results highlighted in red and bold are the best results.The underlined one is the second-best result.}
\vskip 0.15in
\begin{center}
\begin{small}
\begin{sc}
  % 表格内容
  \centering
  \begin{tabular}{l|cc|cc|cc|cc|cc}
    \toprule
    \multicolumn{1}{l|}{Model}&\multicolumn{2}{c|}{PETNN}&\multicolumn{2}{c|}{RNN}&\multicolumn{2}{c|}{LSTM}&\multicolumn{2}{c|}{GRU}&\multicolumn{2}{c}{Transformer}\\
    \cmidrule{2-11}
    \multicolumn{1}{l|}{Metric}&\multicolumn{1}{c}{MSE}&\multicolumn{1}{c|}{MAE}&\multicolumn{1}{c}{MSE}&\multicolumn{1}{c|}{MAE}&\multicolumn{1}{c}{MSE}&\multicolumn{1}{c|}{MAE}&\multicolumn{1}{c}{MSE}&\multicolumn{1}{c|}{MAE}&\multicolumn{1}{c}{MSE}&\multicolumn{1}{c}{MAE}\\
    \hline
    ETTm1      &\textcolor{red}{\textbf{{0.52}}}&\textcolor{red}{\textbf{{0.47}}}&1.54&0.98&1.32&0.86&1.37&0.86&\underline{0.85}&\underline{0.68}\\
    ETTm2      &\textcolor{red}{\textbf{{0.30}}}&\textcolor{red}{\textbf{{0.34}}}&2.80&1.36&2.39&1.18&2.26&1.10&\underline{1.38}&\underline{0.81}\\
    ETTh1      &\textcolor{red}{\textbf{{0.48}}}&\textcolor{red}{\textbf{{0.47}}}&1.54&0.99&1.19&0.82&1.32&0.83&\underline{0.89}&\underline{0.74}\\
    ETTh2      &\textcolor{red}{\textbf{{0.43}}}&\textcolor{red}{\textbf{{0.43}}}&4.87&1.64&3.09&1.35&3.43&1.49&\underline{2.35}&\underline{1.30}\\
    Electricity&\textcolor{red}{\textbf{{0.19}}}&\textcolor{red}{\textbf{{0.29}}}&0.64&0.60&0.56&0.55&0.54&0.54&\underline{0.32}&\underline{0.42}\\
    Traffic    &\textcolor{red}{\textbf{{0.61}}}&\textcolor{red}{\textbf{{0.33}}}&1.46&0.85&1.01&0.54&1.03&0.54&\underline{0.68}&\underline{0.38}\\
    Weather    &\textcolor{red}{\textbf{{0.27}}}&\textcolor{red}{\textbf{{0.28}}}&0.80&0.61&0.44&0.45&0.69&0.58&\underline{0.34}&\underline{0.47}\\
    Exchange   &\textcolor{red}{\textbf{{0.47}}}&\textcolor{red}{\textbf{{0.47}}}&2.85&1.47&2.11&1.22&2.08&1.23&\underline{1.35}&\underline{0.93}\\
    \bottomrule
  \end{tabular}
   
  \label{time series prediction task in basic models}
\end{sc}
\end{small}
\end{center}
\vskip -0.1in
\end{table*}

\begin{table*}[t]
\caption{Long-term forecasting tasks on SOTA methods. All the others are the same as above.}
\vskip 0.15in
\begin{center}
\begin{small}
\begin{sc}
    \begin{tabular}{l|cc|cc|cc|cc|cc|cc|cc}
    \toprule
    \multicolumn{1}{l|}{Model}&\multicolumn{2}{c|}{\textbf{PETNN}}&\multicolumn{2}{c|}{TimesNet}&\multicolumn{2}{c|}{Informer}&\multicolumn{2}{c|}{FEDformer}&\multicolumn{2}{c|}{Mamba}&\multicolumn{2}{c|}{ETSformer}&\multicolumn{2}{c}{DLinear}\\
    \cmidrule{2-15}
    \multicolumn{1}{l|}{Metric}&\multicolumn{1}{c}{MSE}&\multicolumn{1}{c|}{MAE}&\multicolumn{1}{c}{MSE}&\multicolumn{1}{c|}{MAE}&\multicolumn{1}{c}{MSE}&\multicolumn{1}{c|}{MAE}&\multicolumn{1}{c}{MSE}&\multicolumn{1}{c|}{MAE}&\multicolumn{1}{c}{MSE}&\multicolumn{1}{c|}{MAE}&\multicolumn{1}{c}{MSE}&\multicolumn{1}{c|}{MAE}&\multicolumn{1}{c}{MSE}&\multicolumn{1}{c}{MAE}\\
    \hline
     ETTm1  &{0.52}&{0.47}&\textcolor{red}{\textbf{0.40}}&\textcolor{red}{\textbf{0.40}}&0.96&0.73&0.45&0.45&0.47&0.45&0.43&0.42&\underline{0.40}&\underline{0.41}\\
     ETTm2  &\underline{0.30}&\underline{0.34}&\textcolor{red}{\textbf{0.29}}&\textcolor{red}{\textbf{0.33}}&1.41&0.81&0.31&0.35&0.33&0.36&0.29&0.34&0.35&0.40\\
     ETTh1  &{0.48}&{0.47}&\underline{0.46}&\textcolor{red}{\textbf{0.45}}&1.04&0.80&\textcolor{red}{\textbf{0.44}}&\underline{0.46}&0.52&0.50&0.54&0.51&{0.46}&{0.45}\\
     ETTh2  &{0.43}&\textcolor{red}{\textbf{0.43}}&\underline{0.41}&\underline{0.44}&4.43&1.73&0.44&0.45&0.42&0.44&0.44&0.45&0.56&0.52\\
     Electricity &\textcolor{red}{\textbf{{0.19}}}&\textcolor{red}{\textbf{{0.29}}}&\underline{0.20}&\underline{0.30}&0.31&0.40&0.21&0.33&0.22&0.32&0.21&0.32&0.21&0.30\\
     Traffic &\textcolor{red}{\textbf{0.61}}&\textcolor{red}{\textbf{{0.33}}}&0.62&\underline{0.34}&0.76&0.41&{0.61}&0.38&0.78&0.44&0.62&0.40&0.63&0.38\\
     Weather & \textcolor{red}{\textbf{0.25}}&\textcolor{red}{\textbf{{0.28}}}&{0.26}&\underline{0.29}&0.63&0.55&0.31&0.36&0.29&0.31&0.27&0.33&0.27&0.32\\
     Exchange &{0.47}&{0.47}&0.42&0.44&1.55&1.00&0.52&0.50&0.73&0.44&\underline{0.41}&\underline{0.43}&\textcolor{red}{\textbf{0.35}}&\textcolor{red}{\textbf{0.42}}\\
    \bottomrule
    \end{tabular}
    
    \label{time series prediction task with sota methods}
\end{sc}
\end{small}
\end{center}
\vskip -0.1in
\end{table*}

\subsection{Time Series Forecasting Task}

\paragraph{Setup}Time series forecasting is the typical sequence learning  holds significant importance across a multitude of fields, including weather prediction and energy consumption planning. To thoroughly assess the model on time series data, we employed the long-term forecasting benchmark as utilized in TimesNet~\cite{timesnet}, including five application scenarios in actual fields such as ETT~\cite{informer}, Electricity~\cite{electricity}, Traffic~\cite{trafficdata}, Weather~\cite{weatherdata}, and Exchange~\cite{exchange} and conducted experiments on  different length of sequences. 
Utilizing the robust framework provided by the Time Series Library (TSLib)~\cite{timesnet}, we conducted a comprehensive comparison of two distinct categories of baseline. The first comprise traditional foundational models, including RNN, LSTM, GRU, and Transformer. The second is SOTA methods, including TimesNet, Informer, FEDformer, Mamba~\cite{mamba}, ETSformer, and DLinear~\cite{dlinear}.

% The training procedures were meticulously followed according to the parameters specified in the original papers, and we achieved the training results reported therein. 
%% 放appendix configuration details
%For PETN training, due to certain issues, we did not employ the token embedding from the original paper. Instead, we utilized only temporal embedding and position embedding. All other training parameters remained consistent with those used in the other models.

\paragraph{Results}
As shown in Tables \ref{time series prediction task in basic models} and \ref{time series prediction task with sota methods}, PETNN demonstrates superior performance in both traditional models and state-of-the-art (SOTA) methods. Specifically, PETNN reduces MSE and MAE by an average of 60\% compared to Transformer-based methods. When compared with SOTA models, PETNN consistently ranks among the top performers. Despite the heterogeneity and noise in real-world time series data from various domains, our model still demonstrates outstanding performance, particularly in the four tasks among these tasks. This robust performance validates the effectiveness of our novel cross-field theory approach. Due to an unknown issue, unfortunately, during the experiments on PETNN and Mamba, the project's architecture faced problems: Nan. So we discard one of the embeddings. Additionally, we conducted experiments to investigate whether the embedding is effective for other models. The results demonstrated an average decrease of 0.03 in each metric. Furthermore, we urge caution in overemphasizing rankings, as this may lead to overfitting noise. Instead, we advocate for focusing on the intrinsic performance of the model.

%The PETNN model demonstrates exceptional performance across both traditional baseline and SOTA baselines, as detailed in Table\ref{time series prediction task in basic models} and Table\ref{time series prediction task with sota methods}. Specifically, PETNN attains the highest rank among basic models.Within all tasks, PETNN achieved a 60\% reduction on average in both MSE and MAE compared to the Transformer. Moreover, it achieves the best or second-best performance among SOTA methods in most cases. Despite the heterogeneity of the time series data sourced from diverse areas, our model exhibits superior performance, outperforming advanced MLP-based, CNN-based, and Transformer-based models. This robust performance underscores the model's versatility and potential for broad applicability across various domains.

\subsection{Text Sentiment Classification Task}
\paragraph{Setup} Similarly, a sentence is also a type of sequential data, and we aim to test the model through such tasks to validate its ability to understand textual information. We selected the widely-used ACL-IMDB sentiment classification dataset as our benchmark. This dataset consists of movie reviews from the Internet Movie Database (IMDB), labeled as positive or negative, making it a standard benchmark for binary sentiment classification tasks. By testing on this dataset, we evaluate the model's performance in accurately classifying sentiment from text data.

To enable a fair comparison, PETNN was evaluated against several traditional baseline models, including TextCNN~\cite{textcnn}, LSTM, GRU, and MLP. A standardized evaluation framework was employed to ensure consistency across experiments. Specifically, all input sequences were padded or truncated to a fixed length of 300 tokens. Word embeddings were generated using GloVe~\cite{glove}, resulting in 300-dimensional vectors for each word. Consequently, the input to the models was represented as a feature matrix with a shape of $\left[300, 300 \right]$.

\paragraph{Results}

\begin{table}[t]
\caption{Text Sentiment Classification Task.}
\vskip 0.15in
\begin{center}
\begin{small}
\begin{sc}
    \begin{tabular}{c|ccccc}
    \toprule
    Model&PETNN&TextCNN&LSTM&GRU&MLP\\
    \midrule
    Acc. (\%)&\textcolor{red}{\textbf{89}}&\underline{84}&83&81&72\\
    \bottomrule
    \end{tabular}
    \label{text sentiment classification task}
\end{sc}
\end{small}
\end{center}
\vskip -0.1in
\end{table}

The results of our comparative experiments, shown in Table~\ref{text sentiment classification task}, indicate that PETNN significantly outperforms other methods in terms of accuracy. Despite the challenges of the ACL-IMDB dataset, which contains lengthy reviews and presents difficulties in capturing long-term dependencies, PETNN demonstrates exceptional proficiency. By leveraging advanced mechanisms to capture contextual nuances, the model outperforms RNN-based and TextCNN models, effectively utilizing the intrinsic characteristics of text for improved classification performance. Additionally, we do some ablation study on the resistance to interference in the next section and hyperparameter optimization in Appendix C.

\begin{figure}[h!]
    \centering
    \includegraphics[width=\columnwidth]{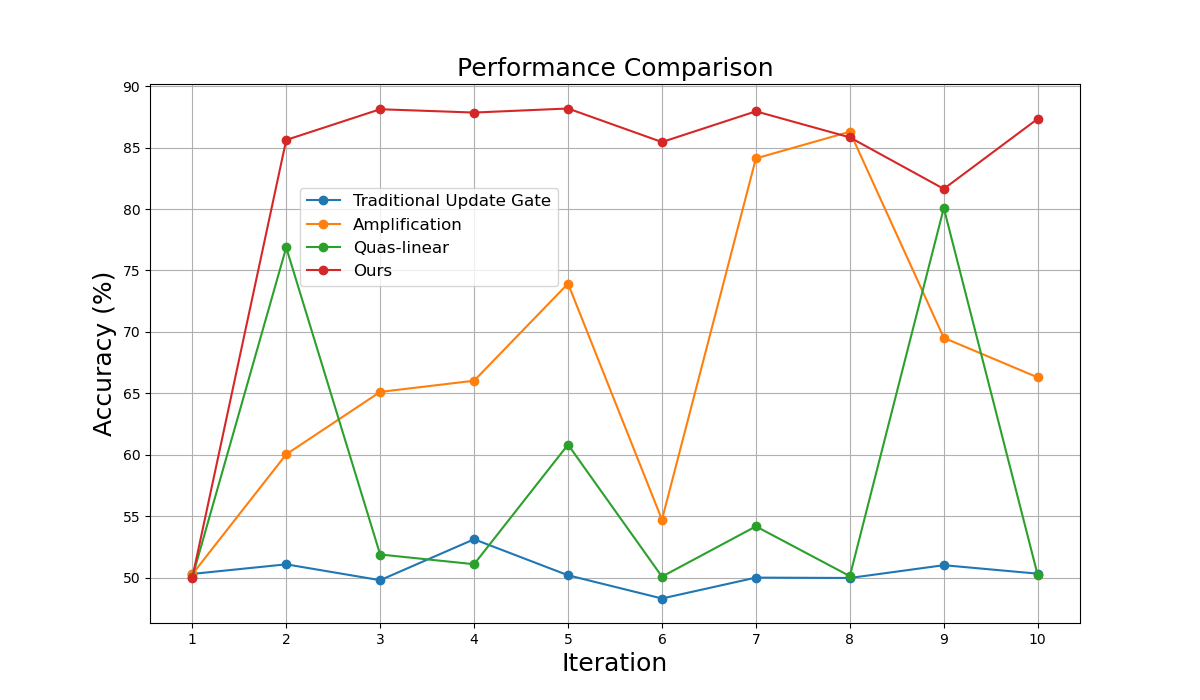} % Reduce the figure size so that it is slightly narrower than the column. Don't use precise values for figure width.This setup will avoid overfull boxes.
    \caption{Comparision of Different Update Methods.}
    \label{Ablation study on updating component}
\end{figure}

%%%%%%%%%%%%%%%%%%%%%%%%%%%%%%%%%%%%%%%%%%%%%%%%%%%%%%%%%%%%%

\section{Ablation Study}

%In this section, we focus on three core aspects of the model: the effectiveness of the information updating mechanism, and the effect of neuronal continuity and independence. 
In this section, we focus on several ablation studies of our model to investigate why it works, including the information updating mechanism, Computational Efficiency analysis, robustness analysis, and neuronal continuity and independence analysis.

%thereby ensuring high efficiency and reliability across diverse input scenarios, and effectively mitigating the issue of long-term dependency

\paragraph{Information Updating Method}

In the PETNN model, the information update mechanism is a core component. We compare it with the commonly used method as following:

\begin{itemize}
    \item \textbf{Traditional Gating Method}: Similar to the gating mechanism in LSTM, this method updates the state $S_t$ in PETNN using forget and update gates.
    \item \textbf{Quasi-linear Transformation Method}: Inspired by linear transformations, the state is updated according to the equation: 
    $S_t = Z_w \cdot S_{t-1} + h_{state}$.
    \item \textbf{Exponential Gating Amplification Method}: Based on xLSTM~\cite{XLSTM}, we introduce an exponential gate (exp gate) to amplify memory information.
    \item \textbf{Self-selective Information Mixing Method}: Drawing inspiration from neuroscience, we propose a novel update mechanism that allows neurons to autonomously select and update information partially.
\end{itemize}

The results, based on the ACL-IMDB dataset (Figure~\ref{Ablation study on updating component}), highlight the performance differences among the four information updating methods. The traditional gating method shows significant performance fluctuations, failing to adapt well to the PETNN model. The quasi-linear transformation method is stable but lacks the accuracy needed, indicating it cannot fully capture the model's complexity. The Exponential Gating Amplification Method performs well initially but declines over time, suggesting issues with overfitting or reliance on early states. In contrast, the Self-selective Information Mixing Method maintains high accuracy and stability throughout training, proving to be the most reliable and effective approach. These results provide strong evidence for the effectiveness of our method and suggest that it offers a clear advantage in optimizing the PETNN model’s performance.

\paragraph{Computational Efficiency Analysis}

Our primary goal is to identify a more efficient alternative to traditional RNN models and Transformer models. While our model may not outperform state-of-the-art (SOTA) models in every task, as shown in Table~\ref{time series prediction task with sota methods}, it offers significant advantages in terms of computational efficiency, as highlighted in Table~\ref{params}. 
\begin{table}[t]
\caption{Efficiency Evaluation in Time Series Forecasting.}
\vskip 0.15in
\begin{center}
\begin{small}
\begin{sc}
\begin{tabular}{lll}
\toprule
Model & FLOPs (M) & Params (M) \\
\midrule
PETNN       & \textcolor{red}{\textbf{170.2133}}   & \textcolor{red}{\textbf{0.045207}}  \\
Mamba\_Simple& 517.767168 & 0.225664 \\
TimesNet    & 18792.9231 & 0.605479 \\ 
Informer    & 34977.1530 & 11.328007 \\
FEDformer   & 38042.8830 & 10.535943 \\
Transformer & 1188.5445  & 10.54     \\
Reformer    & 35597.0580 & 5.794     \\

\bottomrule
\end{tabular}
\label{params}
\end{sc}
\end{small}
\end{center}
%\vskip -0.1in
\label{FLOP}
\end{table}
These efficiency gains are particularly important for real-time applications and scenarios with limited computational resources, PETNN can provide robust performance while maintaining lower computational overhead.

\paragraph{Robustness Evaluation}

To evaluate the stability of PETNN and its ability to alleviate long-term dependency issues, we selected a comment from the ACL-IMDB dataset for testing. To better observe the effect, we artificially reconstructed this comment and divided it into several components, as shown in Table~\ref{visualization}. These components include negative items (from the original review), irrelevant items (repeated nonsense words), and interfering items (positive words). Additionally, Hidden state visualizations and detailed text context can befound in Appendix D.

\begin{table}[t]
\caption{Robustness of PETNN and LSTM on Noisy Data} 
\vskip 0.15in
\begin{center}
\begin{small}
\begin{sc}
    \begin{tabular}{p{4cm}|c|c}
    \toprule
    \multirow{2}{*}{Text context}& \multicolumn{2}{c}{Model}  \\
    \cmidrule{2-3}
    %\midrule
     &    PETNN     & LSTM \\
    \midrule
    \textbf{(Neutral Item:)} The characters receive less emphasis overall...best.& Postive & Postive \\
    \midrule
    \textbf{(Negative Item:)} This movie utterly disappoints ...  I just didn't care about any of the characters. & Negative & Negative \\ 
    \midrule
    \textbf{(Negative Item:)} This movie utterly disappoints. ...  I just did not care about any of the characters.  \textbf{(Neutral Item:)} The characters receive less emphasis overall...to be the best.& Negative &  Positive \\
    \midrule
    \textbf{(Negative Item:)} This movie utterly disappoints. ...  I just did not care about any of the characters. \textbf{(Irrelevant Item:)} Lorem ... ipsum \textbf{(Interfering Items:)} \textcolor{red}{This is  ... fantastic.}\textbf{(Irrelevant Item:)} Blah... blah.\textbf{(Neutral Item:)} The characters receive less emphasis overall...to be the best. & \textcolor{red}{Negative} & \textcolor{red}{Postive} \\
    \bottomrule
    \end{tabular}
\label{visualization}
\end{sc}
\end{small}
\end{center}
\vskip -0.1in
\end{table}

As shown in the Table~\ref{visualization}, both PETNN and LSTM predict a positive sentiment when faced with a neutral item, and they both predict negative sentiment when presented with a negative item. However, when a negative item precedes a long neutral context, PETNN effectively captures the negative sentiment at the beginning. This demonstrates PETNN's ability to alleviate long-term dependency problems. When irrelevant items are introduced, LSTM fails to filter out the noise, leading to a positive sentiment prediction. In contrast, PETNN retains its robustness by correctly identifying and maintaining the negative sentiment throughout, ultimately delivering an accurate negative prediction. 

\paragraph{Neuronal Continuity and Independence}

In this section, we explore whether neurons need to be reset at each time step, considering the implications of neuronal continuity and independence. Our aim is to investigate how this could potentially enhance the modeling effectiveness of neural networks and the information updating mechanism.

\begin{table}[t]
\caption{Neuronal Continuity and Independence Analysis.} 
\vskip 0.15in
\begin{center}
\begin{small}
\begin{sc}
    \begin{tabular}{c|cccc}
    \toprule
    (Time, Energy) &(1,1)&(0,1)&(1,0)&(0,0)  \\
    \midrule
    Accuracy&\textcolor{red}{\textbf{88.13\%}}&87.61\%&87.56\%&87.92\%\\
    \bottomrule
    \end{tabular}
\label{Continuity and Inheritability}
\end{sc}
\end{small}
\end{center}
\vskip -0.1in
\end{table}

Where time and energy represent whether $T_t$ and $C_t$ are retained (1) or reset (0), as shown in Table~\ref{Continuity and Inheritability}.The results show that simultaneous reset and retention achieve better results among the four schemes, specially, retention $(1,1)$ achieving the highest effect $88.13\%$. Consequently, keep continuity in both time and energy which can enhance the information process. This insight has the potential to guide the development of more effective neural network and significantly enhance our understanding of memory mechanisms. 

% The issue of neuronal continuity and independence is particularly complex and critical in neuroscience. At the center of this issue is the question of whether neurons need to undergo some reset before receiving new inputs, which is directly related to the learning and memory mechanisms of neural networks. This section will focus on this issue and analyze it from the perspective of neuronal continuity and independence, with the aim of exploring its potential enhancement of modeling effectiveness.
\section{Conclusion}

In this paper, we propose the Physics-inspired Energy Transition Neural Network (PETNN), a novel approach that leverages material properties to construct neural networks from the fundamental perspective of constituent particles.

Our experiments show that PETNN not only outperforms traditional models but also competes with state-of-the-art (SOTA) methods. We also explore why PETNN works effectively, focusing on the impact of its physics-inspired architecture in addressing long-term dependency issues in sequence modeling.

We believe PETNN offers significant insights for advancing sequence model research. Going forward, our goal is to evolve PETNN into a more generalized architecture, using its physics-based foundation to further enhance performance and extend its applicability across various domains.

%In this paper, we propose ``Physics-inspired Energy Transition Neural Network"(PETNN), a pioneering approach that harnesses material properties to construct neural networks from the fundamental perspective of constituent particles.

%Through different types of experiments, we verified that PETNN is not only significantly better than the traditional basic models but also competes SOTA methods. To address the long-term dependency problem faced by traditional sequence models, we proposed a novel solution and achieved SOTA results.

% We believe that PETNN provides significant insights for advancing sequence model research. Moving forward, our primary objective is to evolve PETNN into a more generalized architecture, leveraging its physics-inspired foundation to enhance performance and expand its applicability across a wider range of domains. 

\section*{Impact Statement}
This paper presents work whose goal is to advance the field of Machine Learning. There are many potential societal consequences of our work, none which we feel must be specifically highlighted here.

\bibliography{example_paper}
\bibliographystyle{icml2025}

%%%%%%%%%%%%%%%%%%%%%%%%%%%%%%%%%%%%%%%%%%%%%%%%%%%%%%%%%%%%%%%%%%%%%%%%%%%%%%%
%%%%%%%%%%%%%%%%%%%%%%%%%%%%%%%%%%%%%%%%%%%%%%%%%%%%%%%%%%%%%%%%%%%%%%%%%%%%%%%
% APPENDIX
%%%%%%%%%%%%%%%%%%%%%%%%%%%%%%%%%%%%%%%%%%%%%%%%%%%%%%%%%%%%%%%%%%%%%%%%%%%%%%%
%%%%%%%%%%%%%%%%%%%%%%%%%%%%%%%%%%%%%%%%%%%%%%%%%%%%%%%%%%%%%%%%%%%%%%%%%%%%%%%
\newpage
\appendix
\onecolumn
\section*{Appendix}

\subsection*{Appendix A: Physics Background of Energy Transition Model}
\textbf{Energy Level.} Energy levels are discrete quantities of energy that electrons in an atom can possess. They are typically determined by the principal quantum number $n$ in quantum mechanics. 
In quantum mechanics, the energy levels are related to atomic type and electronic status, showing complex mathematical form. Here, we take the hydrogen-like atom as instance, energy level of which can be represented by

\[
E_n = - \frac{Z^2 e^4 m_e}{8 \epsilon_0^2 h^2 n^2}
\]

where:
\begin{itemize}
    \item \( E_n \) is the energy of the level \( n \),
    \item \( Z \) is the atomic number,
    \item \( e \) is the elementary charge,
    \item \( m_e \) is the electron mass,
    \item \( \epsilon_0 \) is the permittivity of free space,
    \item \( h \) is Planck's constant,
    \item \( n \) is the principal quantum number (positive integer).
\end{itemize}

The negative sign indicates that the energy is lower than the zero energy level, which is the energy of a free electron.

\textbf{Energy Level Transition.} Energy Level Transitions
When an electron transitions between different energy levels, it either absorbs or emits a photon. The energy of this photon corresponds to the difference between the initial and final energy levels. The formula for the energy change is shown below.

Energy level transitions occur when an electron moves between two discrete energy levels in an atom. The energy difference \( \Delta E \) between the initial state with quantum number \( n_i \) and the final state with quantum number \( n_f \) is given by:

\[
\Delta E = E_{n_i} - E_{n_f}
\]

Substituting the expression for \( E_n \), we get:

\[
\Delta E = - \frac{Z^2 e^4 m_e}{8 \epsilon_0^2 h^2 n_i^2} + \frac{Z^2 e^4 m_e}{8 \epsilon_0^2 h^2 n_f^2}
\]

\[
\Delta E = \frac{Z^2 e^4 m_e}{8 \epsilon_0^2 h^2} \left( \frac{1}{n_f^2} - \frac{1}{n_i^2} \right)
\]

This energy difference \( \Delta E \) corresponds to the energy of the photon \( E_{\text{photon}} \) emitted or absorbed during the transition:

\[
E_{\text{photon}} = \Delta E
\]

where \( E_{\text{photon}} \) is the energy of the photon associated with the transition.
\newpage
\subsection*{Appendix B: Pseudocode} 

\begin{algorithm}[]
    \caption{Physics-inspired Energy Transition Neural Network.}
    \begin{algorithmic}
        \REQUIRE $X_t, [T_{t-1}, C_{t-1}, S_{t-1}]$
        \ENSURE  $[T_t,C_t,S_t]$
        \STATE {-----INITIALIZATION-----}
        \STATE $Z_t \gets W_{Z_t} [X_t,S_{t-1}] + b_{Z_t}$
        \STATE $Z_c \gets W_{Z_c} [X_t,S_{t-1}] + b_{Z_c}$
        \STATE $Z_w \gets W_{Z_w} [X_t,S_{t-1}] + b_{Z_w}$
        \STATE $I_t \gets W_{I_t} X_t + b_{I_t}$
        \STATE $R_t \gets W_{R_t} X_t + b_{I_t}$
        \STATE {--------UPDATE----------}
        \STATE $T_t \gets R_t*(T_{t-1}+Z_t)-1$
        \IF {$T_t \leq 0$}
        \STATE $T_t \gets 0,m \gets 1$
        \STATE $C_t \gets (1-m) * C_{t-1} + I_t + Z_c$
        \ELSE
        \STATE $m \gets 0$
        \STATE $C_t \gets (1-m) * C_{t-1} + Z_c$
        \ENDIF
        \STATE $h_{t} \gets f(S_{t-1},(1-m)\cdot C_{t-1},X_t)$
        \STATE $S_t \gets \sigma((1-Z_w)S_{t-1}+Z_w h_{t}))$

    \end{algorithmic}
    \label{Algorithm}

\end{algorithm}

\subsection*{Appendix C: 
Experimental configuration and supplementary experiments}

\paragraph{Experimental configuration}
We provide the dataset descriptions and experiment configurations in Table~\ref{Basic information of time series data}. All experiments are repeated three times, implemented in PyTorch and conducted on a single NVIDIA V100 12GB GPU.

The detailed settings for the tasks are presented as follows:
\begin{itemize}
    \item {\textbf{Time Series Forecasting Task}}: This task uses the composite multi-dimensional time series datasets including ETT, electricity, traffic, weather and exchange. We follow standard protocol and split all datasets into training, validation and test sets in chronological order by the ratio of 6:2:2 for the ETT dataset and 7:1:2 for the others.
    
    \item{\textbf{Text Sentiment Classification Task}}: This task uses the ACL-IMDB dataset which is a collection of 50,000 movie reviews from the Internet Movie Database (IMDB) for binary classification tasks. Among them, 25000 movie reviewers for training and 25000 for testing. Every review is truncated into the fixed-length sequence of 300, and GloVe.6B.300d~\cite{glove} is used to convert every word into a vector representation.
    % in dimension 300
    \item {\textbf{Image Classification Task}}: This task uses the MNIST dataset~\cite{mnist}, which contains 60,000 training samples and 10,000 test samples. These samples are images of handwritten digits from 0 to 9, each image being 28$\times$28 pixels. We process these images into 784-dimensional feature vectors and keep the original dim for sequence data.
\end{itemize}

\begin{table*}[htbp]
  % 表格内容
\vskip 0.15in
\begin{center}
\begin{small}
\begin{sc}
 \caption{Dataset descriptions. All the experiments use the ADAM optimizer.}
    \begin{tabular}{l|c|c|c|c}
    \toprule
    Datasets&Dim&Series Length&Dataset size&Information(Frequency)\\
    \midrule 
    MNIST&28&\{28,764\}&(60000, 10000)&Handwritten digits\\    
    \midrule
    ACL-IMDB&300&\{300,600\}&(25000, 25000)&Movie Review\\
    \midrule
    ETTm1,ETTm2& 7 &\{96,192,336,720\}&(34465, 11521, 11521)&Electricity(15 mins)\\
    \midrule 
    ETTh1,ETTh2&7 &\{96, 192, 336, 720\} &(8545, 2881, 2881)&Electricity(15 mins)\\
    \midrule 
     Electricity&321&\{96, 192, 336, 720\}&(18317, 2633, 5261)&Electricity(Hourly)\\
    \midrule 
     Trafﬁc&862&\{96, 192, 336, 720\}&(12185, 1757, 3509)&Transportation(Hourly)\\
    \midrule 
     Weather&21&\{96, 192, 336, 720\}&(36792, 5271, 10540)&Weather(10 mins)\\
    \midrule 
     Exchange&8&\{96, 192, 336, 720\}&(5120, 665, 1422)&Exchange rate(Daily)\\
    \bottomrule
  \end{tabular}
  
  \label{Basic information of time series data}
\end{sc}
\end{small}
\end{center}
\vskip -0.1in
\end{table*}

More experimental configurations are as follows:
\begin{itemize}
    \item Time Series Forecasting Task: All the experiments are conducted under the framework of TSLib. All the baselines that we reproduced are implemented based on configurations of the original paper or official code. For PETNN, we construct it with the structure: input-\{cell dim=64\}-output.
    \item Text Sentiment Classification Task: As we describe above, every review is processed into the 300 $\times$ 300 feature matrix. We built PETNN with the structure: input-\{cell dim=64\}-output. For Textcnn, we build the constructor: input-\{Conv2d(2)-Relu-MaxPool2d\}-\{Conv2d(3)-Relu-MaxPool2d\}-\{Conv2d(4)-Relu-MaxPool2d\}-output. For LSTM and GRU, their hidden dimension are all set at 256. For FC, we use the structure: input-linear(300$\times$300, 128)-output.
    \item Image Classification Task: To ensure the validity and fairness of comparisons, we selected models with specific layer configurations for our experiments. We chose a single layer and two layers of fully connected layer: linear(784,256), linear(784,784)-linear(784,256), a 3-layer KAN with structure: input-\{28 $\times$28, 64, 10\}, a single-directional LSTM with, a 1-layer CNN with Conv2d(3,3), and PETNN with cell dim set to 64 as our basic units.
\end{itemize}

For metrics, we adopt the mean square error (MSE) and mean absolute error (MAE) for time series forecasting task. For text sentiment classification task and image classification task, we use the accuracy to evaluate the performance. For some activation functions, we also tried to maintain consistency as much as possible. To ensure the stability of the results as much as possible, we adopted zero initialization.

\paragraph{Image Classification Task}
Fundamentally, image data can be viewed as a type of sequential data, with the primary challenge being how to unfold it and determine appropriate times-teps and features. In this task, we aim to thoroughly evaluate PETNN’s performance in processing data from different perspective, highlighting its adaptability for more general tasks. Consequently, this work lays the foundation for extending PETNN to more general neural network architectures in future research. To achieve this, we employ two distinct data processing strategies: 

To preserve the original image size, we treat the image width as the number of channels and the columns as the content for each timestep, converting it into time-series-like data. As our preferred conversion method, we compared it with several architectural models, including LSTM, a representative RNN-based model; Convolutional Neural Network (CNN)\cite{cnn}, which is well-suited for image processing tasks; and KAN\cite{kan}, a recently proposed foundational model.

For the alternative method, each image is flattened into a one-dimensional vector, a technique commonly employed in MLP-based models. We compared the PETNN model with single-layer and two-layer fully connected networks (FC)~\cite{MLP}, using corresponding layers. This comparison allowed us to evaluate PETNN’s performance against traditional fully connected architectures, providing insights into its ability to handle image data in a sequential manner.
\begin{table}[ht]
\caption{Image classification Task on different scenarios. }
\vskip 0.15in
\begin{center}
\begin{small}
\begin{sc}
  \begin{tabular}{lccc}
    \toprule
    \multicolumn{1}{l|}{Model}&\multicolumn{1}{c|}{Layer}&\multicolumn{1}{c|}{Input size}&\multicolumn{1}{c}{Accuracy}\\
    \toprule
    \multicolumn{1}{l|}{FC}&\multicolumn{1}{c|}{1}&\multicolumn{1}{c|}{[1,784]}&\multicolumn{1}{c}{90.91\%}\\
    \multicolumn{1}{l|}{PETNN}&\multicolumn{1}{c|}{1}&\multicolumn{1}{c|}{[1,784]}&\multicolumn{1}{c}{\textcolor{red}{\textbf{96.80\%}}}  \\
    \midrule
    \multicolumn{1}{l|}{FC}&\multicolumn{1}{c|}{2}&\multicolumn{1}{c|}{[1,784]}&\multicolumn{1}{c}{96.91\%}\\
    \multicolumn{1}{l|}{PETNN}&\multicolumn{1}{c|}{2}&\multicolumn{1}{c|}{[1,784]}&\multicolumn{1}{c}{\textcolor{red}{\textbf{97.20\%}}} \\
    \midrule
    \multicolumn{1}{l|}{KAN}&\multicolumn{1}{c|}{3}&\multicolumn{1}{c|}{[28,28]}&\multicolumn{1}{c}{96.70\%}\\
    \multicolumn{1}{l|}{LSTM}&\multicolumn{1}{c|}{1}&\multicolumn{1}{c|}{[28,28]}&\multicolumn{1}{c}{98.07\%}\\
    \multicolumn{1}{l|}{CNN}&\multicolumn{1}{c|}{1}&\multicolumn{1}{c|}{[28,28]}&\multicolumn{1}{c}{\underline{98.47\%}}\\
    \multicolumn{1}{l|}{PETNN}&\multicolumn{1}{c|}{1}&\multicolumn{1}{c|}{[28,28]}&\multicolumn{1}{c}{\textcolor{red}{\textbf{99.03\%}}}\\
    \bottomrule
  \end{tabular}

  \label{simple image classification task}
\end{sc}
\end{small}
\end{center}
\vskip -0.1in
\end{table}

The results of the image classification task involving PETNN and various baseline models are detailed in Table~\ref{simple image classification task}. Notably, PETNN achieved the highest rankings in both experimental sections, underscoring its remarkable superiority over other models such as FC, LSTM, CNN, and KAN. This performance highlights PETNN's exceptional adaptability across diverse data processing context, making foundation for more general tasks.
\paragraph{Hyperparameter Optimization}

\begin{table}[ht!]
\caption{Hyperparameter Optimization for PETNN.}
\vskip 0.15in
\begin{center}
\begin{small}
\begin{sc}
  \begin{tabular}{c|c|c|c}
    \toprule
    \multicolumn{1}{c}{length}&\multicolumn{1}{c}{Batch size}&\multicolumn{1}{c}{Cell dim}&\multicolumn{1}{c}{Accuracy}\\
    \hline
    \multirow{8}{*}{300}& \multirow{2}{*}{32} & 32  &88.18\% \\
    \cmidrule{3-4}
    &  & 64 &87.08\% \\
    \cmidrule{2-4}
    & \multirow{2}{*}{64} & 32 &\textcolor{red}{\textbf{88.44\%}} \\
    \cmidrule{3-4}
    &  & 64 &86.77\% \\
    \cmidrule{2-4}

    & \multirow{2}{*}{128}& 32 &\underline{88.20\%}\\
    \cmidrule{3-4}
    &  & 64 &87.02\% \\
    \cmidrule{1-4}
    \multirow{8}{*}{600} & \multirow{2}{*}{32} & 32  &\textcolor{red}{\textbf{89.14\%}} \\
    \cmidrule{3-4}
    & & 64  &85.59\% \\
    \cmidrule{2-4}
    & \multirow{2}{*}{64} & 32  &\underline{89.11\%}\\
    \cmidrule{3-4}
    & &  64  &85.18\%\\
    \cmidrule{2-4}
    & \multirow{2}{*}{128}& 32  &88.92\%\\
    \cmidrule{3-4}
    & &  64  &85.81\%\\
    \bottomrule
  \end{tabular}
  
  \label{hyperparameters evaluation}
\end{sc}
\end{small}
\end{center}
\vskip -0.1in
\end{table}

We conducted a comprehensive parameter optimization study on the PETNN model, with a focus on evaluating its performance across sequences of varying lengths and the dimensions of its internal memory modules. Specifically, we performed detailed comparisons using sequences of lengths 300 and 600, with internal memory unit dimensions set at 32 and 64, respectively. The test results, as depicted in Table~\ref{hyperparameters evaluation}, clearly illustrate the model's performance under different configurations. Consequently, we find that the model achieved optimal results when configured with an input sequence length of 600, a batch size of 32, and a neuron dimension of 32. These findings not only substantiate the model's superior performance in handling longer sequences but also provide optimized parameter settings for practical deployment.

\subsection*{Appendix D: Showcases}

\begin{figure}[h!]
    \centering
    \includegraphics[width=0.8\linewidth]{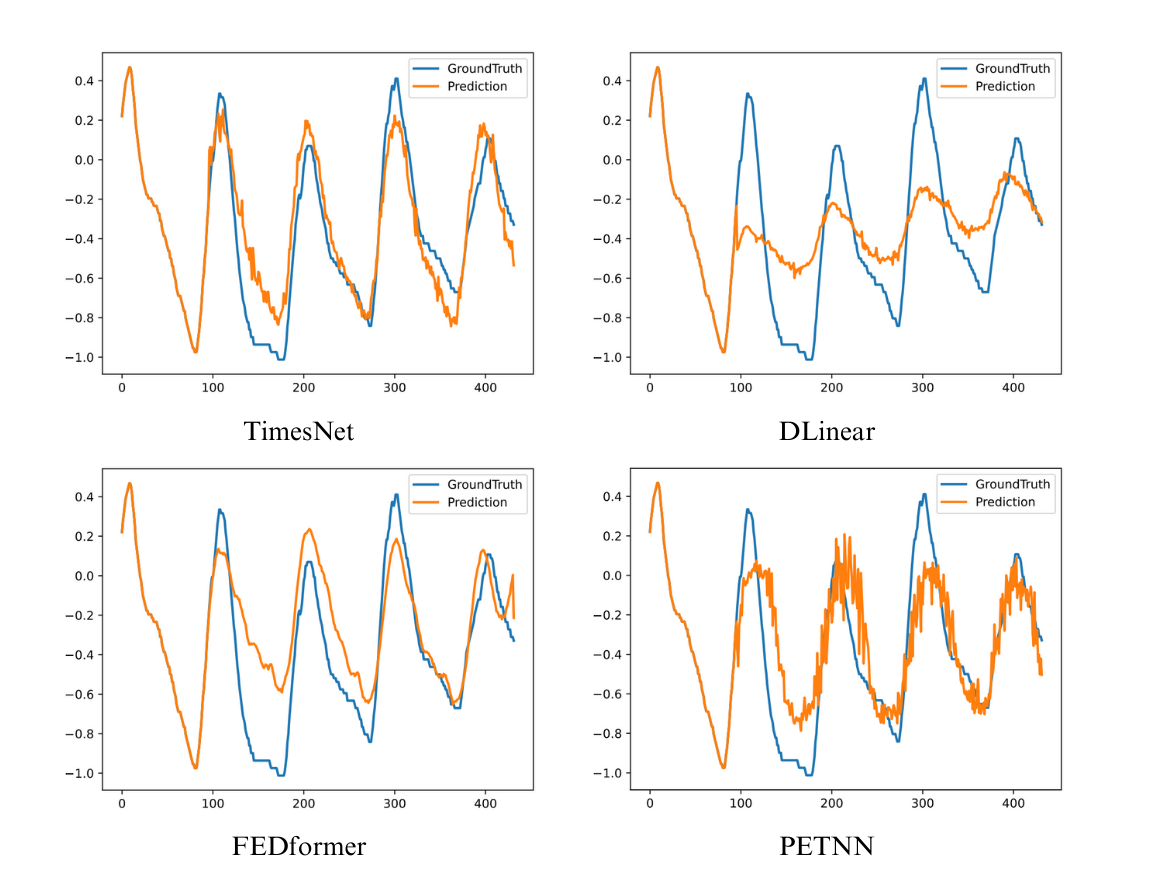}
    \caption{Visualization of ETTm2 prediction by different models under the input-96-predict-336 setting.The blue lines stand for the ground truth and the orange lines stand for the prediction.}
    \label{showcases}
\end{figure}

\paragraph{Showcase of time series forecasting}
To provide a clear comparison among different models, we provide some showcases of the prediction task in the time series forecasting task. We choose the ETTm2 datasets as an example, shown in Figure~\ref{showcases}.

In this task, the MLP-based models degenerate a lot, it could only follow the overall trend of the data, but could not capture the detailed fluctuations. In comparison, PETNN performed better than other models to some extent. Although PETNN still lacks sensitivity in some turning points, it maintains a more consistent fit with the real data overall. PETNN can track the actual data patterns more accurately and shows obvious advantages in tasks that require precise matching of real data.

\begin{figure}[ht!]
\vskip 0.2in
\begin{center}
\centerline{\includegraphics[width=0.6\columnwidth]{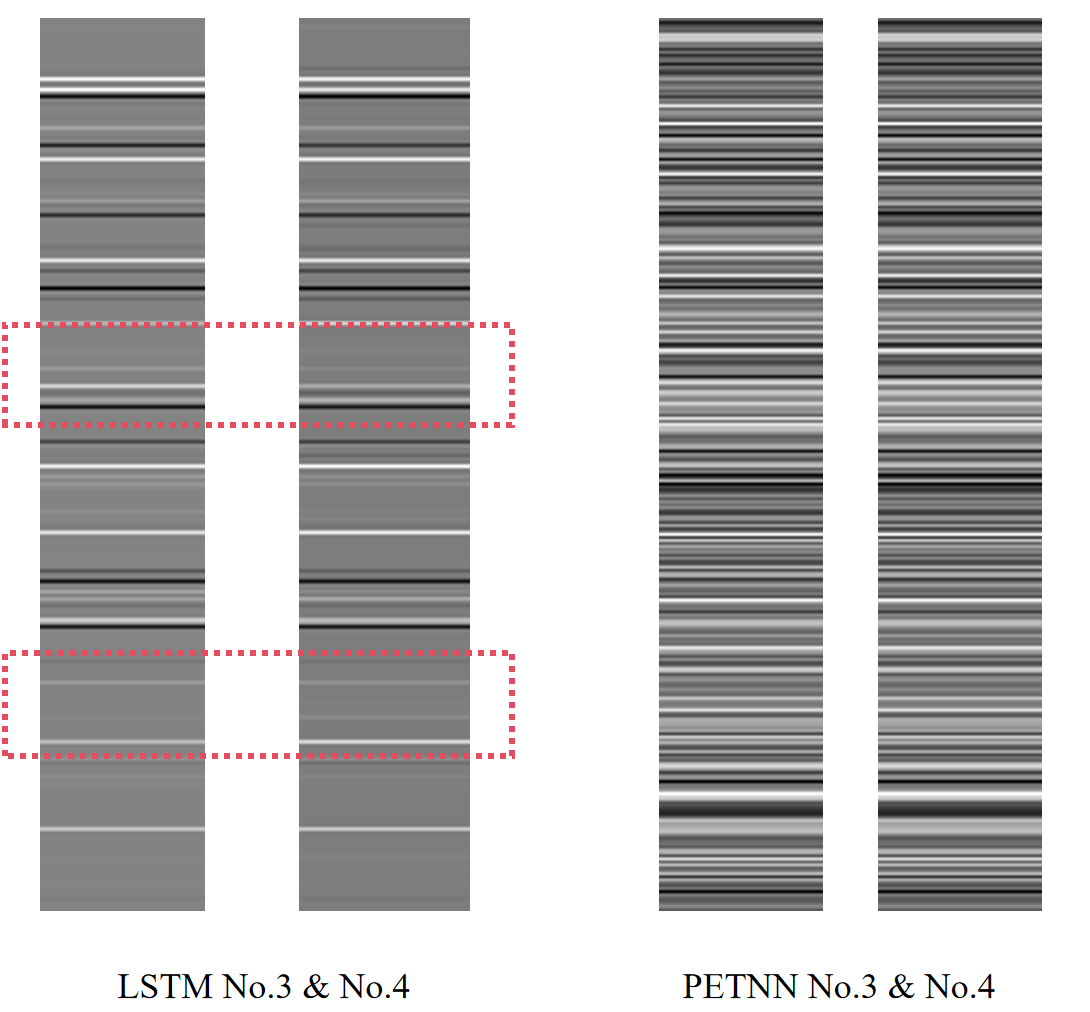}}
\caption{Visualization of hidden state in last step}
\label{cnl}
\end{center}
\vskip -0.2in
\end{figure}

\paragraph{Visualization of hidden state}
In this part, we mainly show the noisy case in Table~\ref{exam} and the visualization of the hidden state in last step Figure~\ref{cnl}.

\begin{table}[ht]
\caption{Test Noisy Case for Robustness Evaluations} 
\vskip 0.15in
\begin{center}
\begin{small}
\begin{sc}
    \begin{tabular}{p{13cm}}
    This movie utterly disappoints from the very beginning. Aside from the terrific sea rescue sequences, of which there are very few, I just did not care about any of the characters. Blah blah blah blah blah blah. lorem ipsum lorem ipsum lorem ipsum lorem ipsum This is  fantastic fantastic. Blah blah blah blah blah.Blah blah blah blah blah blah blah blah blah blah blah blah blah blah.The characters receive less emphasis overall. The character we should focus on is Ashton Kutcher, who interacts with Costner. Eventually, when we are well past the halfway point, Costner provides some insight into Kutcher's background. We learn why Kutcher is driven to be the best.
    \end{tabular}
\label{exam}
\end{sc}
\end{small}
\end{center}
\vskip -0.1in
\end{table}

From the visualization, we can find that facing the different sentences whether it have some irrelevant items in it. We can find that LSTM show minor different. However, PETNN stay the same which show the robustness of the PETNN.

%%%%%%%%%%%%%%%%%%%%%%%%%%%%%%%%%%%%%%%%%%%%%%%%%%%%%%%%%%%%%%%%%%%%%%%%%%%%%%%
%%%%%%%%%%%%%%%%%%%%%%%%%%%%%%%%%%%%%%%%%%%%%%%%%%%%%%%%%%%%%%%%%%%%%%%%%%%%%%%

\end{document}